# Experiments using Belief Functions and Weights of Evidence incorporating Statistical Data and Expert Opinions

by

Mary McLeish, Paulyn Yao, Mathew Cecile and
Tatianya Stirtzinger (DVM)

University of Guelph
Guelph, Ontario, N1G 2W1
Canada

## Abstract

This paper presents some ideas and results of using uncertainty management methods in the presence of data in preference to other statistical and machine learning methods. A medical domain is used as a test-bed with data available from a large hospital database system which collects symptom and outcome information about patients. Data is often missing, of many variable types and sample sizes for particular outcomes is not large. Uncertainty management methods are useful for such domains and have the added advantage of allowing for expert modification of belief values originally obtained from data. Methodological considerations for using belief functions on statistical data are dealt with in some detail. Expert opinions are incorporated at various levels of the project development and results are reported on an application to liver disease diagnosis. Recent results contrasting the use of weights of evidence and logistic regression on another medical domain are also presented.

* This work is supported by NSERC operating grant #A4515. The author is currently on leave at the MRC Biostatistics Unit, Cambridge, England.

## 1. Introduction

Hospitals have had database systems in place for many years, usually handling billing and admissions procedures primarily. More recently systems are being installed (eg. University of British Columbia teaching hospital) which collect on-line results of many medical tests. Companies such as Epic Systems Corporation of Madison, Wisconsin are producing medical information systems to handle this type of data. A study of this data collected over time should prove valuable for the enhancement of medical expert systems. Although many epidemiological studies are carried out on medical data, often the data used does not represent exactly the tests carried out at a particular hospital and upon which immediate diagnoses are made. However, as the data collected by these systems is dependent on the type and number of cases coming to that particular hospital, sample sizes might not be large, especially for some outcomes (eventually one would hope to use a large conglomerate of data collected from a number of sites, provided there could be a standardization of tests). The question arises of how to properly analyze this type of data and many methods immediately present themselves - statistical models (regression, discriminant analysis etc.), machine learning techniques (pattern matching, various cluster analysis methods, tree building etc.) and methods from uncertainty management, although the latter have been more often used with information obtained from expert opinions. A recent paper by Schwartz, Patil and Szolovits [15] suggests bringing some of the older statistical methods and the analysis of data into the development of medical expert systems.

In the projects undertaken at the University of Guelph a database system is available which handles the usual admission and billing procedures. The system also stores a considerable amount of medical information on each patient, including bacteriology, clinical pathology, parisitology, radiology and other patient information such as age, sex, breed, presenting complaint, treatment procedures, diagnosis and outcome. In clinical pathology, much of the data is electronically generated by the lab equipment. A distributed ORACLE DBMS running on a 10 processor (386) SEQUENT parallel machine manages the data. A considerable amount of outcome data is available with autopsies always being conducted and extensive follow-up work. This provides an excellent test-bed for ideas and indeed a number of the methods mentioned have been experimented with for equine abdominal problems in work reported by Cecile [3] and McLeish [10]. This preliminary study showed many of the difficulties of working with data which often contained missing values, had many variable types, was not necessarily well-behaved (fitting models or distributions well), and involving a great many input parameters (70 in one domain and about 100 in the other). Sample sizes for special populations (very young horses for example), for which the behaviour of variables can be different from the general population, were small. Techniques which found the most "significant" variables and based the entire diagnosis completely on this, were of no use when the information for one of these variables was missing. In earlier studies by Ducharme et al [5], a multiple stepwise discriminant analysis in a recursive partition model was used to determine a decision protocol. The significant parameters were found to be abdominal pain, distension and to a lesser extent, the color of abdominal fluid. Another problem with this method, which puts all the diagnostic weight on these few variables, concerns the fact that they are very subjectively assessed and other more accurately determined variables have been dropped. In our previous work [3, 10],an evidence combination scheme which does not delete variables is proposed (based on the work of I. Good and incorporating a method of dealing with dependencies) and compared with a number of other methods. A short summary of recent results of this study is given in Section 2 of this paper. This work had considered and tried a variety of machine learning methods, coming to the conclusion that they could supplement other methods but, in themselves, were not as effective as the evidence combination method. Work is continuing on this domain with the experts examining the weights found for symptoms and symptom groups from the data and are attempting to modify them - especially when small sample sizes have been involved.

The new domain concerns the diagnosis of liver diseases in small animals on the basis of pathology data. This domain has a very large number of outcomes (originally around 30) which were pruned using expert opinions to a group of 14 sets. The statistical data samples were not very large per outcome going back over about 3 years of collected cases) and were smaller per outcome than that used for the equine domain. It was decided to apply Dempster-Shafer theory to this domain, initially extracting information from the data and then modifying it with expert opinions (the experts themselves preferred to proceed in this direction). Although some ideas of how to derive support functions from statistical evidence are given in [14], it is also stated there that the suggestions presented "are not implied by the general theory of evidence exposited in the preceeding chapters" and that they "must be regarded as "conventions" for establishing degrees of support, that can be justified only by their intuitive appeal and by their success in dealing with particular examples".

A number of methods have been tried and some initial results are available. Section 3 describes this work in detail.

2. Weights of Evidence Approach

The problem of diagnosing surgical cases in horses with abdominal pain is a very difficult one and has been the result of many large studies and much attention in the world of veterinary medicine eg. Ducharme [5]. The diagnositic process begins with the owner or local veterinarian who must make a preliminary decision on whether or not to ship the animal to the Guelph hospital where large animal surgery can be performed. At least in the first phase, we have focused on the second stage of diagnosis when the patient arrives at the hospital (a referral population).

Time, which is a significant factor in the decision for shipping (duration
of pain etc.) is now critical because there is usually not enough of it. Often
decisions are made before all test results are taken (accounting for some of our
missing data) and it is the speed with which decisions must be made which has
been a factor in the need for an automated system. The clinicians and other
specialists recorded symptoms on a diagnostic chart. These values were later
entered into the database system. A decision was made on the basis of a
weighting scheme. The expert defined weights did not vary with the strength of
the symptom. The information was simply tallied up by hand on these sheets and
used, at least as a significant guideline, in making a final decision.
(Variables such as pain, pulse rate, temperature were not recorded as varying
with time on these sheets).

Our first system consisted of an automated version of these charts with
expert defined weights and a simple combination scheme. It was implemented in
QNAIL (Lisp/Apl) on a Microvax II. However, its performance on a set of new
cases was often inconclusive and generally below the 'experts' own success
records. We decided to exploit the data on past cases to improve the weighting
scheme, while imitating the over-all strategy. This would also involve using
weights which were a function of symptom strengths (instead of the constant
weights).

**Description of Methodology and Summary of Results.**

A weight of evidence approach originally proposed by Turing and later by
Good [6,7] and Minsky [12] was adopted with some modifications. In order to
assign weights to different levels of symptom strengths, experts were requested
to draw fuzzy membership functions [17]. From these, Yager's notion of the
probability of a fuzzy event [16] was used to incorporate results into the
model. In particular, the weight of evidence is defined as

$$W(H:E) = \log \frac{p(E/H)}{p(E/\overline{H})} \quad \text{or} \quad W(H:E) = \log \frac{O(H/E)}{O(H)} \quad \text{where } O(H) \text{ represents the odds of } H.$$

In order to deal with regions of usually low, normal and high ranges of
continuous variables, physicians were asked to provide membership functions
(sometimes overlapping) indicating the belief in the outcomes.

Yager proposes that the probability of a fuzzy event be a fuzzy subset
(fuzzy probability:

$$P(A) = \bigcup_{\alpha=0}^{1} \alpha \left[ \frac{1}{P(A_\alpha)} \right]$$

where $\alpha$ specifies the $\alpha$-level subset of A and since $P(A_\alpha) \in [0,1]$, $P(A)$ is a
fuzzy subset of $[0,1]$. This fuzzy subset then provides a probability of A for
every $\alpha$-level subset of A. Thus, depending on the required (or desired) degree
of satisfaction, a probability of the fuzzy event A is available. In our case
the desired level of truth is that which maximizes the bias of this event to the
hypothesis. We define this optimal $\alpha$-level to be:

$$\underset{\alpha}{\text{Max}} \quad |W(H:E_\alpha)| \quad \alpha \in [0,1]$$

$W(H:E_\alpha)$ is the weight of evidence of the strong $\alpha$-level subset $E_\alpha$ provided
towards the hypothesis H. The $\alpha$-level which maximizes the bias of a fuzzy enent
to a hypothesis (or null hypothesis) is the optimal $\alpha$-level for minimizing
systematic noise in the event.

Attempts were made to account for higher order dependencies by considering symptom groups. The computational complexity of searching for, in some sense, the 'best' symptom group is extremely high. This can be reduced by not considering sets when the frequency of occurrence is too small to be relevant. The symptom groups are ranked using a weighted average of two factors: reliability (lower confidence interval of $p(H|E)$) and specificity (related to the size of the symptom groups). Two heuristic search methods were adopted to attempt to find a 'best' symptom group covering for each new case being diagnosed. Version 1 uses chi-square tests of significance to ensure two-way and three-way independence among all the pieces of evidence. Details are as follows:

1. Find the 'best' (highest ranked piece of evidence).

2. Eliminate all other pieces of evidence not independent of this one.

3. Find the next best piece of evidence.

4. Eliminate all other pieces of evidence not independent of this one or any possible pairwise combination of evidence applied containing this one. Repeat 3 and 4 until no evidence is left to be applied.

The second version adopts a simpler approach which does not check for independence but merely enforces a maximal overlapping condition (set at 1) upon the evidence. The 'best' pieces of evidence are chosen iteratively as before. Below are results based on 68 new cases (300 cases were used for the training set). The input variables consisted of 25 clinical symptoms, further refined into a set of about 70 using the ranges for some variables obtained from the 'fuzzy' method and expert opinions. Shown is a standard ROC curve (Receiver operating characteristic). The decision criteria are 0.2, 0.4, 0.6 and 0.8. Here a true positive is a correctly diagnosed surgical candidate. Version 1 out-performs version 2 at several points. It is interesting to note that an earlier form of the first method, which checked only pairwise independence, was out-performed by the second method. It is interesting to note that the clinician's own success rate runs around 70% overall.

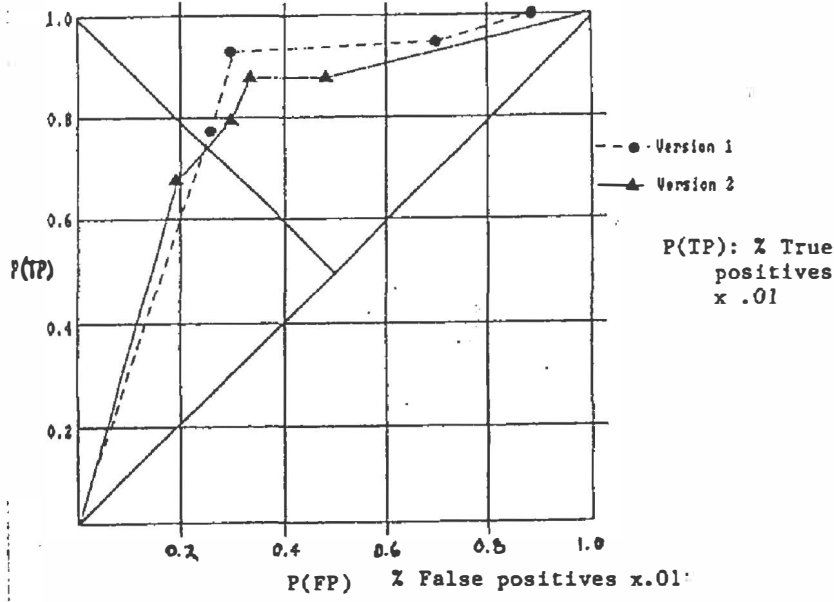

Logistic regression was run on the same data set. Here a regression model of the form
$$y = \log \frac{p}{1-p} = \beta_0 + \beta_2 X_1 + \ldots \beta_k X_k$$
is used relating each of the independent variables to the log odd's ratio. More information can be found in [11]. The probability of a response is estimated by

256

a back transformation. The clinical data set was appropriately revised, splitting several nominal variables into new variables with binary outcomes. A 95% significance level was used. The same training set and new cases were tested. Problems with missing data for the key (significant) parameters in the regression model reduced the test data set to 38. Below is a ROC curve for the results.

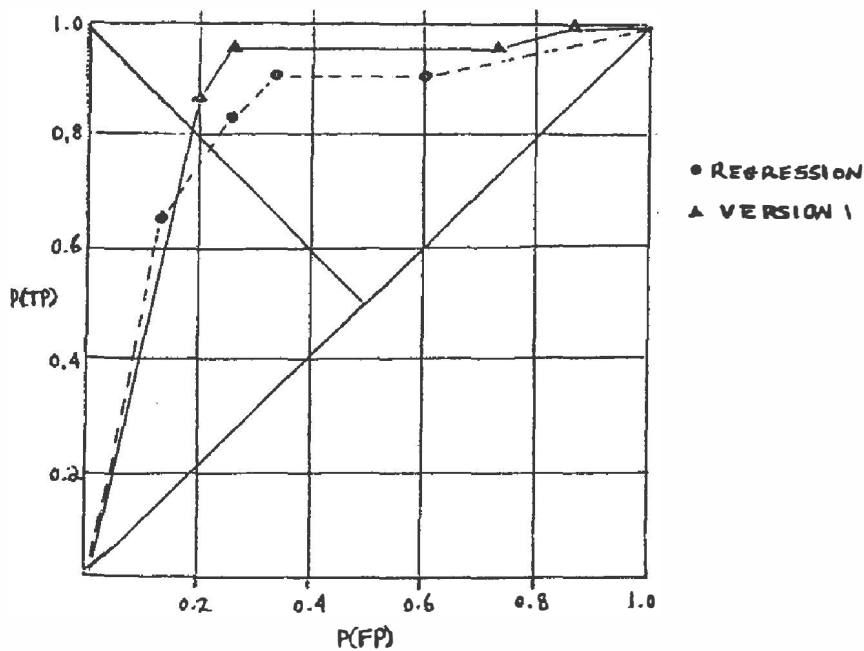

The evidence combination scheme thus performed better than regression on the new data. It more closely resembles the clinician's own approach and results can be presented to them in a form which indicates the most significant symptoms and symptom groups for a given patient. The diagnosis is not based on only a few 'significant' variables (as is the case with many machine learning and other statistical methods). It is thus less sensitive to missing data at the diagnostic stage. No additional data has been lost due to missing values initially either, as is the case with regression models. It has the added advantage that the weights can be altered by expert opinions, especially for rarer symptoms involving small sample sizes. This work is now in progress.

The implementational work was carried out on a SEQUENT parallel machine with 4 INTEL 80386's. The method was coded in C and Pascal and made considerable use of the programming interface to the ORACLE RDBMS. This provided a powerful blend of procedural and non-procedural languages in a parallel programming environment. Further details of the computational work including the use of a blackboard architecture and a C-ORACLE combination for fuzzy relations can be found in [2,9].

### 3. An Application of Dempster-Shafer Theory using Statistical Data and Expert Opinions.

#### 3.1 Use on Statistical Data:

Use on Statistical Data

Another domain under investigation concerns the use of pathology data in the diagnosis of liver disease in small animals. A number of considerations motivated the development of an automated system for this domain.

(1) The relevant data is available in the database system at the hospital and is generated on-line as results are produced by the biochemistry analyzers.
(2) The expert's success rate has not been particularly good in this domain, partly due to the large number of input parameters and outcome variables involved.
(3) The demands on the few specialists often leaves interns and other students with considerable responsibilities. An aid to them both as a learning tool and a diagnostic aid would be very valuable.

The interconnections between the variables in this domain are not well understood and attempts at producing a network of causal relations resulted in a very complicated structure of doubtful accuracy. The outcome space consistes of 30 different types of liver disease and the laboratory tests produced 19 biochemical parameters and 21 hematologic parameters (all continuous variables, eg. albumins, calcium, glucose, cholesterol, total protein, red blood cell, lymphocytes, reticulocytes etc.). Initially we had only 150 cases and solicited expert opinions to reduce the outcome space to set of 14 classes based on the similarity of the lesions and the pathogenesis of liver diseases. Examples of these classes are Primary and Metastic Tumours, Hepatocellular Necrosis, Hepatic Fibrosis and Cirrhosis, Hepatic Congestion, Hepatic Atrophy and Hypoplasia etc. Even between these 14 classes, the data sample is not large enough to be convincingly discriminatory. As our intention was to sollicit expert opinions to modify our statistically determined results and these were also not always well defined on singleton classes, it was decided to use the belief function approach from Dempster-Shafer theory. (In the equine domain, subsets of the small outcome space were meaningless and subtantially more data per outcome was available). The problem formulation does not lend itself to the hierarchical or network approaches of Zarley et al [18] or Lauritzen and Spiegelhalter [8]. Although tests of correlations between parameters were carried out, detailed hierarchies were not well known.

## Methodology

In D-S theory, the frame of discernment $\Theta$ is not the set over which a probability measure is defined. The power set, $2^\Theta$, is the basic set upon which judgements are made. The subtle distinction between a probability mass assigned to a singleton and a larger subset containing that singleton is not immediately arrived at by considering frequency of occurrence information. If different outcomes were very well discriminated over some range of a continuous variable (i.e., normal range for some medical symptom), values could be easily assigned. Unfortunately plots of our 40 symptoms in the liver disease domain revealed this not to be the case (even with outcomes considered as members of $2^\Theta$). The range of continuous variables to be considered as individual symptoms were specified by the experts. These regions could be further refined to produce a greater degree of discrimination, but the computational complexity of the domain was already so high that we decided to keep to the expert defined ranges.

## Method 1

There are several ideas which present themselves of how to make use of data to assign mass and belief functions in such situations. One idea comes from [14], where a support function $S_x(A)$ is defined by the following for all non-empty $\bar{A}$:

$$S_x(A) = 1 - Pl_x(\bar{A}),$$

where the plausibility function $Pl_x(A) = \dfrac{\max\limits_{\theta_i \in A} q_{\theta_i}(x)}{\max\limits_{\theta_i \in \Theta} q_{\theta_i}(x)}$ and $\bar{A}$ stands for A complement.

Here $\theta_i$ is an outcome and x is a symptom. The functions $\{q_{\theta_i}\}_{\theta_i \in \Theta}$ is a statistical specification which obeys the rule that x renders $\theta_i \in \Theta$ more plausible that $\theta_i'$ whenever $q_{\theta_i}(x) > q_{\theta_i'}(x)$. A possible choice for $\{q_{\theta_i}\}$ are likelihood functions, see [14].

One may work out what the associated mass functions are. (This is necessary for the method of implementation).

**Theorem 3.1**

If $f_x(\theta_i)$ represents the frequency of occurrence of outcome $\theta_i$ given symptom x and these outcomes are sorted in descending order by the f function for each x, then the non-zero mass functions are given by the formulae:

$$m_x(\theta_1) = \frac{f_x(\theta_1) - f_x(\theta_2)}{f_x(\theta_1)}$$

$$m_x(\theta_1,\ldots,\theta_j) = \frac{f_x(\theta_j) - f_x(\theta_{j+1})}{f_x(\theta_1)}$$

$$m_x(\theta_1,\ldots,\theta_n) = \frac{f_x(\theta_n)}{f_x(\theta_1)}$$

where $\theta = \{\theta_1,\ldots,\theta_n\}$ and $f_x(\theta_i) \geq f_x(\theta_j)$ $i \leq j$.

**Proof:** The consonnance of the belief function resulting from the formula for $S_x(A)$ is noted in Theorem 11.1 in [14]. Thus its focal elements can be arranged in order so that each one can be contained in the following one. It is not difficult to determine that the foci are the sets given above (see also [14]).

To determine the belief values consider $\theta_1$.

$$Bel\{\theta_1\} = 1 - \frac{f_x(\theta_2)}{f_x(\theta_1)} = m_x(\theta_1) \text{ by definition of } S_x(A).$$

Consider now $Bel\{\theta_1,\theta_2\} = 1 - \frac{f_x(\theta_3)}{f_x(\theta_1)} - Bel\{\theta_1\}$.

Therefore $m_x\{\theta_1,\theta_2\} \neq 0$ and equals $Bel\{\theta_1,\theta_2\} - m_x(\theta_1) = \frac{f_x(\theta_2) - f_x(\theta_3)}{f_x(\theta_1)}$.

In general $Bel\{\theta_1,\ldots,\theta_j\} = 1 - \frac{f_x(\theta_{j+1})}{f(\theta_1)}$

and $m\{\theta_1,\ldots,\theta_j\} = Bel\{\theta_1,\ldots,\theta_j\} - Bel\{\theta_1,\ldots\theta_{j+1}\}$ (due to the consonnance)

$$= \frac{f_x(\theta_j) - f_x(\theta_{j+1})}{f_x(\theta_1)}.$$

Now Support $(\phi) = 0$ and under the assumption that the m values sum to 1,

$$m_x(\theta) = \frac{f_x(\theta_n)}{f_x(\theta_1)}.$$

This result was the form used in the implementation, which was written to accommodate general mass function assignments (not necessarily support functions). As indicated in the introduction, however, there is no real theoretical justification for this method and Shafer himself suggests experimentation to determine an appropriate method for a particular application. Some other methods are presented below.

**Methods 2:**

A search for a simple support function for each symptom was carried out by looking for individual symptoms or sets of symptoms with likelihoods greater than 0.5. As the values were based on frequency data, only one singleton could have such a high value. If none have a suitably high value, then sets of size two are examined and the pair with highest value greater than 0.5 is chosen. This process is continued to higher sized sets if necessary.

One problem concerns what to do with the remaining mass. Although D-S theory normally would put it on $\theta$, the remaining frequencies of occurrences are on the complement of the support subsets. Two methods were actually tried on the data: 2A: putting the remaining mass on $\bar{A}$ where $m(A) \geq 0.5$ and 2B: assigning it to $\theta$.

An algorithm for main part of method 2A (2B is similar):

* sort all the freq (a), where $A \subseteq \theta$ and A is a singleton, from the highest to the lowest values.
* from the highest to the lowest values, do the following:
  - if freq(A) > 0.5, then
    * $B \leftarrow A$
    * $m(B) \leftarrow freq(A)$
  - else
    while $m(B) \leq 0.5$
    { to get total $m(B) > 0.5$}
    * $B \leftarrow B \cup A$
    * $m(B) \leftarrow m(B) + freq(A)$
    * get the next freq(A)
  - get the next lower freq(A), say freq(X) - if there is any
  - while freq(X) equal to freq(A)
    { to include all the singletons in which their frequencies are the same}
    * $B \leftarrow B \cup X$
    * $m(B) \leftarrow m(B) + freq(X)$
    * $freq(A) \leftarrow freq(X)$
    * get the lower freq(X), and assign it to freq(X)
  - while freq(X) > 0
    { collect all singletons in which their proportions are greater than zero}
    * $C \leftarrow C \cup X$
    * $m(C) \leftarrow m(C) + freq(X)$
    * get the next lower freq(X), and assign it to freq(X)
  - where $B \subseteq \theta$, $A \subset \theta$, $X \subset \theta$, A, X are singletons.

**Methods 3:**

Several variations on a computationally intense method were considered. These assigned mass function values based on observations of frequencies to all subsets and then used a variety of normalization techniques: normalizing over the entire $2^\theta$ or normalizing across subset groups of the same size first and then normalizing at the end. Variations were also produced in which $\theta$ was assigned values of φ or 1 before normalizing. The normalization had the affect of spreading the mass over singleton and larger subsets in such a way that the relative frequencies are preserved. However, there is not really any justification for the amount of weight that is ultimately assigned to pairs etc. instead of singletons. In the absence of data sets that are well discriminated within a symptom region, this method essentially spreads some of the mass, that would have been assigned to singletons in a probability model, out over higher order subsets in a uniform manner. This then always allows for some doubt as to whether the outcome is exactly a singleton.

### 3.2 Results of Domain Experiments

As mentioned earlier the domain consists of the diagnosis of liver diseases in small animals. The full diagnostic process is actually quite complicated in this domain, proceeding through a first phase narrowing down the diagnosis to certain types of liver disease. The number of outcomes was originally 75 which was then reduced to 15 using expert opinions and preprocessing and finally 14 after a particular outcome variable was first predicted. The data consisted of 151 cases presenting themselves at the University of Guelph hospital during the years 1986 to 1988. There were 40 basic symptoms, all observed as continuous variables. Using expert's advice for intervals, these were into 115 discrete variables (mostly ternary, but some were bivariate). Tests were run for pairwise independence between variables and a decision level requiring that the Pearson Correlation coefficient be >0.5 was used together with expert opinions to reduce the original set of 40 symptoms by 11. Results were obtained before and after this reduction.

Of the 15 possible outcomes, one of these (#4) represented a rather imprecise diagnosis, indicating liver disease existed but the exact type was unknown - even after further patient follow-up. Initially the 151 cases were used to determine mass function values and then 21 new cases were tested using methods 1, 2A and 2B. The results of over all diagnostic accuracy for outcome 4 were:

    method 1  : 47.6%
    method 2A : 80.95%
    method 2B : 23 %

The methodologies were run again with only those cases for which a more specific outcome was known. (Actually many runs were made on a variety of data sets caused by the stage at which outcomes were determined, subtle problems with input variables etc., but a representative sample of these is presented).

Total number of cases used for assessing mass function values (outcome 4 dropped) = 89.

Total number of new cases used for testing purposes for the runs: 21.

The runs CD3, CD5, and CD7 refer to the following situations:
- CD3: uses all variables.
- CD5: drops some of the dependent variables based on a combination of using expert opinions and the results of a Pearson correlation test. (The expert changed the results of the statistical analysis in some cases).
- CD7: drops some of the dependent variables according to an algorithm which searches among pairwise dependencies for variables which might be related to more than one node and leaves only one remaining if a cumulative coefficient is high enough.

**Further explanation of symbols**

S: the highest Bel(A) among the singletons corresponds to the 'expected outcome'.
NONS: consider Bel(A) where A is a singleton or non-singleton. There is an outcome in A which is equal to the 'expected' outcome.
F: failure (non-matched).

Performance in %'s: RECENT DATA SET (1)

|  | METHODS 1 | | | METHODS 2A | | | METHODS 2B | | |
|---|---|---|---|---|---|---|---|---|---|
| S | 35.71 | 21.4 | 21.43 | 42.86 | 21.43 | 21.43 | 35.6 | 21.4 | 21.43 |
| NONS | 14.29 | 21.4 | 35.71 | 42.86 | 78.57 | 78.57 | 21.4 | 50.0 | 57.14 |
| F | 50.0 | 57.2 | 42.86 | 14.28 | 00.00 | 0.00 | 42.9 | 28.6 | 21.43 |

LARGER DATA SET GOING BACK TO OLDER RECORDS OF SAME CASES (2)

|  | METHODS 1 | | | METHODS 2A | | | METHODS 2B | | |
|---|---|---|---|---|---|---|---|---|---|
| S | 42.86 | 35.7 | 40.71 | 46.43 | 35.72 | 40.71 | 37.85 | 30.7 | 30.71 |
| NONS | 12.15 | 25.7 | 27.86 | 31.43 | 54.28 | 49.29 | 20.7 | 40.0 | 48.57 |
| F | 45.0 | 38.6 | 31.43 | 22.14 | 10.00 | 13.33 | 41.45 | 29.3 | 20.72 |
|  | CD3 | CD5 | CD7 | CD3 | CD5 | CD7 | CD3 | CD5 | CD7 |

COMPARISON OF TOTAL MATCHED CASES:

|  | METHODS 1 | | | METHODS 2A | | | METHODS 2B | | |
|---|---|---|---|---|---|---|---|---|---|
| DATA (1): | 50.0 | 42.8 | 67.14 | 85.72 | 100.0 | 100.0 | 57.1 | 71.4 | 78.57 |
| DATA (2): | 55.01 | 61.4 | 68.57 | 77.86 | 90.0 | 90.0 | 58.55 | 70.7 | 79.28 |
|  | CD3 | CD5 | CD7 | CD3 | CD5 | CD7 | CD3 | CD5 | CD7 |

Tests for statistically significant differences between methods were run showing a differentiation between methods 2A and the other two methods - with 2A showing the best overall performance. However for singleton sets method, 1 was almost as accurate. It is interesting to observe that methods 2B, method 1 and method 2A placed, in that order, decreasing weight on $\theta$. This accounts for the relative power of these methods to provide information about second level diagnoses. This is also the reason for the even bigger spread in performance on the set containing outcome 4, which tended to dominate the results because of its high frequency of occurrence. Reducing the data set to account for strong dependencies, seemed to improve the total matched cases but not the singleton diagnoses. The 'expert opinion' derived variable selection generally performed worse than using statistical tests on the data. Methods 3 were so computationally intense that they were not feasible for the application. They were tried on a few cases using a reduced symptom input set and showed very promising results.

The pathologist's accurracy for determining liver disease (outcome 4) is 60%, compared with 47.6% (method 1), 80.95%, method 2A and 23% for 2B. The pathologist's success rate at determining a singleton diagnosis is between 15 and 20% for all stages (based on the concensus of a group of doctors). If symptom groups were allowed of fairly small sizes of the type represented by our NONS sets, the success rate would be about 50% (to be compared with the total matched cases in the tables). Thus use of the data dramatically improved predictive power for this domain. It was very interesting to study these assessments, as the data we used were the very cases seen by the doctors over the past 2-3 years. The doctors were making frequency assessments based on these same observations (although not every doctor saw every case) - perhaps mixed with judgements of a more general nature. The differences with the data-derived values were considerable.

A careful study is being made of the combined information and the first set of m-values has been modified to reflect the expert opinions in a few cases. Eleven of the 120 values (approx. 40 symptoms with low, normal and high ranges) were replaced by an 'expert' value. (These were mostly hematology variables.) The runs used the largest data set with combined visits. The methodology corresponds to CD5 in the previous tables.

|  | Method 2A | Method 2B |
| --- | --- | --- |
| S | 68.2 | 36.4 |
| NONS | 17.9 | 40.9 |
| F | 7.1 | 22.7 |
| Total matched cases: | 86.1 | 77.3 |

The prediction of singleton values has improved dramatically for method 2A (from 35.72) and slightly for 2B (from 30.7). However, the grouped diagnosis is slightly lower for 2A (from 90) and is higher for 2B (from 70.7). Further experimentation is taking place here.

The implementation of the Dempster-Shafer methods was in 'C', run on a SEQUENT parallel machine. Results are obtainable in essentially instantaneous real time. The program allows for any type of mass function to be given in order to handle a variety of methods of specification. (This precluded algorithms such as in Barnett [1], where mass is assigned only to singletons.) The program output provides a summary of the case symptom information, all the belief intervals, and then a shorter list of the strongest outcomes (belief $\geq .5$) listing singletons first.

Conclusions

Two medical domains have been studied on which the intention was to build diagnostic systems making considerable use of statistical data. The two domains are similar in that they have a large number of input parameters; however, the first domain had a very small outcome set and the second a large outcome set, especially for the size of the data set involved. Both domains presented problems with missing data, parameters with many possible values and the absence of a well-defined structure of causal relationships between the input variables. The solutions proposed adopt methods which :

1. rely on <u>all</u> the available variables to make a diagnosis

2. allow for subjectively assigned values to replace or modify terms when data samples are small

3. are less sensitive to missing values both in the initial determination of weights or belief and at the diagnostic stage (due in part to 1)

4. were modified to incorporate parameter dependencies

5. would run in real time with an essentially immediate response time

6. used expert opinion at several levels of their development

In both domains, using statistical data produced systems outperforming the experts. The question remains of how to improve performance even further - clearly the training set can be increased with time. However, rarer cases and especially difficult diagnoses might still be hard to properly assess. It is here that a careful modification of the statistically determined information with expert opinions can help. The process is a complicated one of looking for significant differences between the expert's beliefs and those obtained from the data (especially when data samples are small), modifying the beliefs (weights) and then testing to see if the modification actually improves performance. Some initial results in the liver disease domain have shown an improvement.

Further findings include the result that the weights of evidence approach outperformed logistic regression on the first domain. On the second domain, techniques for using Dempster-Shafer theory on statistical data are presented and assessed. This is shown to be a very useful method for the liver disease domain, in which singleton outcomes are particularly difficult to predict.